# Towards Neural Knowledge DNA

Haoxi Zhang [1], Cesar Sanin [2], Edward Szczerbicki [3],

[1] Chengdu University of Information Technology, Chengdu, China

[2] The University of Newcastle, NSW, Australia

[3] Gdansk University of Technology, Gdansk, Poland

```
Haoxi@cuit.edu.cn
Cesar.Sanin@newcastle.edu.au
Edward.Szczerbicki@zie.pg.gda.pl
```

**Abstract:** In this paper, we propose the Neural Knowledge DNA, a framework that tailors the ideas underlying the success of neural networks to the scope of knowledge representation. Knowledge representation is a fundamental field that dedicate to representing information about the world in a form that computer systems can utilize to solve complex tasks. The proposed Neural Knowledge DNA is designed to support discovering, storing, reusing, improving, and sharing knowledge among machines and organisation. It is constructed in a similar fashion of how DNA formed: built up by four essential elements. As the DNA produces phenotypes, the Neural Knowledge DNA carries information and knowledge via its four essential elements, namely, Networks, Experiences, States, and Actions.

**INTRODUCTION**

Knowledge representation is a fundamental field that dedicate to representing information about the world in a form that computer systems can utilize to solve complex tasks (Davis et al. 1993). It is the study of thinking as a computational process. Then, what is knowledge? This is a question that has been discussed by philosophers since the ancient Greeks, and it is still not totally demystified. Drucker P. F. (2011) defines it as "*information that changes something or somebody - either by becoming grounds for actions, or by making an individual (or an institution) capable of different or more effective action*". While the Oxford Dictionary (2015) defines Knowledge as "*facts, information, and skills acquired through experience or education; the theoretical or practical understanding of a subject*". O'Dell and Hubert (2011) claim that Knowledge is not knowledge until the information inside itself has been taken and used by people. And for scientists and researchers in the AI field, we can argue it as "knowledge is not knowledge until the information inside itself has been taken and used by computers, machines, and agents".

Consequently, a good knowledge representation shall easy to be used by different systems to allow storing, reusing, improving, and sharing knowledge among these systems. A survey carried out by Liao (2003) found that there were generally seven categories of knowledge-based technologies and applications developed until 2002. In another study (Matayong & Mahmood 2012), after analysing 30 published articles between 2003 and 2010 from high quality journals, found nine core theories in the knowledge-based area. However, there are limitations to these technologies: most of them are designed for one specific kind of product; they don't have standard knowledge presentation; most systems lack the capability for information sharing and exchange and most of these systems only focus on supporting a particular stage of a product lifecycle (Li, Xie, & Xu 2011).

Recent studies (LeCun et al. 2015; Gallant 2015; Lescroart et al. 2013) in artificial neural networks (ANN) and psychology have found that the image representations in ANN are very similar to those in bio brains; which inspires us that why do not organise and store knowledge as or close to the way how it exists in the human brain?

In this paper, we propose the Neural Knowledge DNA (NK-DNA), a framework adapting ideas underlying the success of neural networks to the scope of knowledge representation for neural network-based knowledge discovering, storing, reusing, improving, and sharing.

**THE NEURAL KNOWLEDGE DNA**

The Neural Knowledge DNA (NK-DNA) is designed to store and represent the knowledge captured by its domain. It is constructed in a similar fashion of how DNA formed (Sinden 1994): built up by four essential elements. As the DNA produces phenotypes, the Neural Knowledge DNA carries information and knowledge via its four essential elements, namely, *States*, *Actions*, *Experiences*, and *Networks* (Figure 1.).

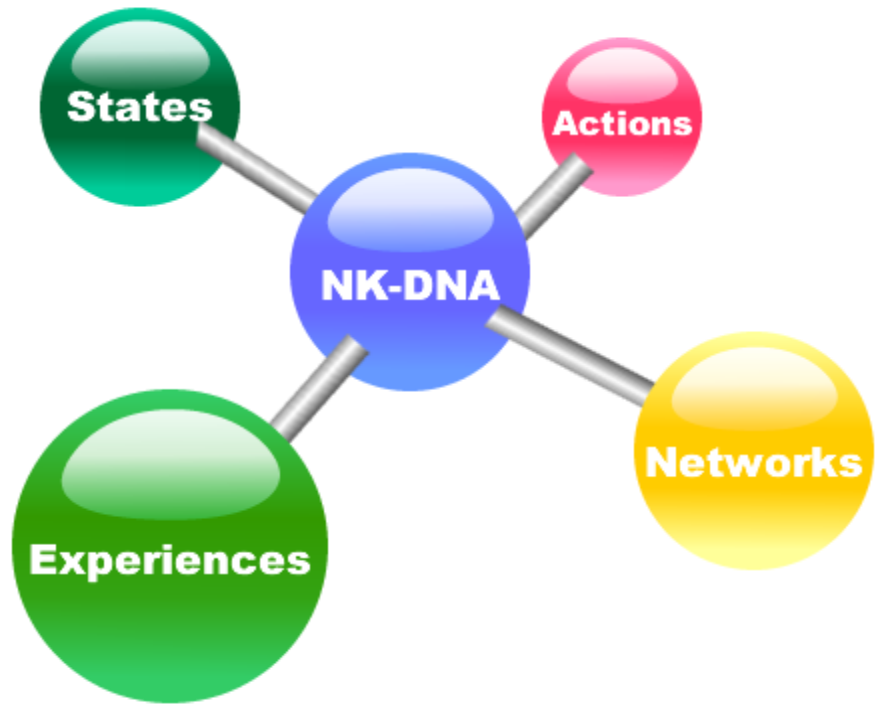

Figure 1. Structure of the NK-DNA

The NK-DNA's four-element combination is also inspired by reinforcement learning and Markov Decision Processes (Sutton & Barto 1998; Michael 2015): ***States*** are situations in which a decision or a motion can be made or performed, ***Actions*** are used to store the decisions or motions the domain can select. While ***Experiences*** are domain's historical operation segments with feedbacks from outcomes, which stored as $e_t = (s_t, a_t, r_t, s_{t+1})$ at each time-step ***t***. And ***Networks*** store the detail of neural networks for training and using such knowledge, such as network structure, deep learning framework used (if a third-party deep learning framework is used, like MxNet, Caffe, etc.), and weights. Figure 2 shows a concept of the NK-DNA-carried knowledge.

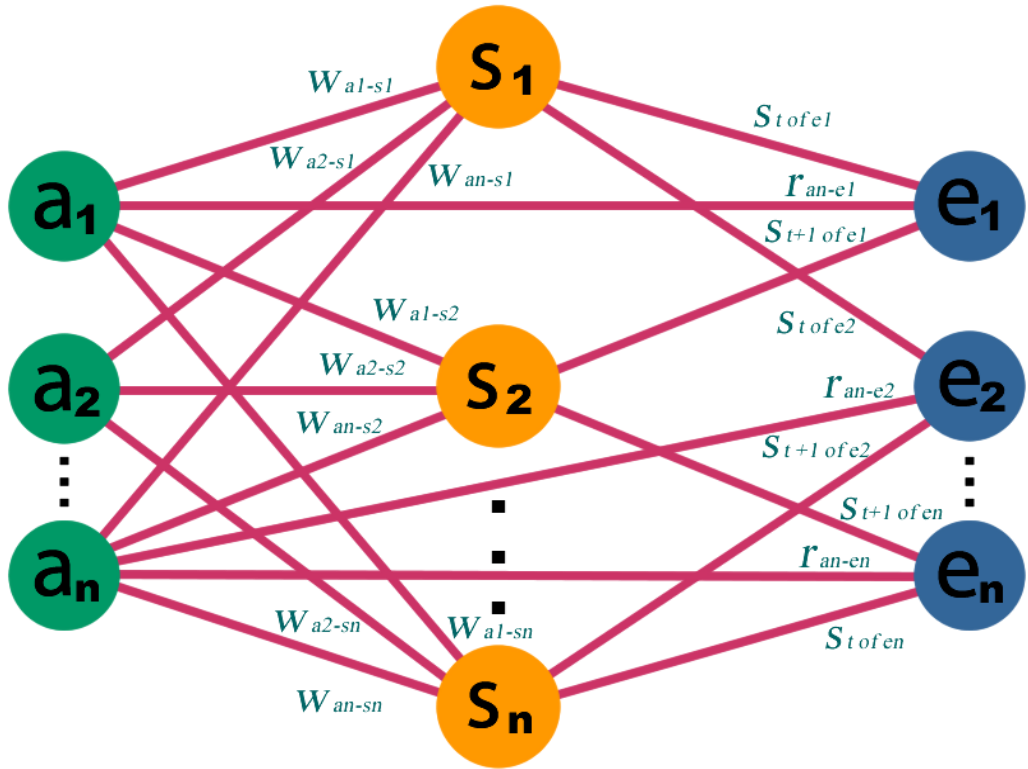

Figure 2. Concept of the NK-DNA-carried knowledge

## INITIAL EXPERIMENT

We examined our NK-DNA in a very simple maze problem. The NK-DNA supported the agent to store the knowledge of the maze (Figure 3.).

In this initial experiment, the agent is expected to find the shortest path to *block 8* from *block 1*. And the agent uses reinforcement learning (Sutton & Barto 1998; Michael 2015) and ANN to learn the maze.

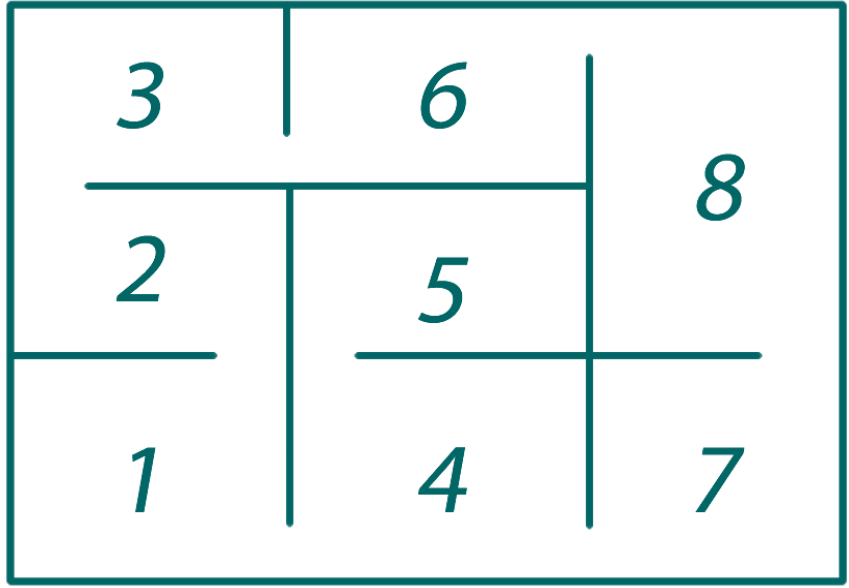

Figure 3.   The very simple maze

First, the *States* and *Actions* are pre-defined, for example, there are 8 states, and for *state 1*, the agent can either "*go to block 2*" or "*go to block 4*" (i.e. the actions of state 1). Then, the agent starts from *state 1*, and randomly picks an action of its current state to explore the maze as long as it reaches the *state 8*, and the agent gets a reward for reaching *state 8*. Meanwhile, the agent stores every single movement (i.e. from one state to another state) with reward from it as an *Experience* ($s_t$, $a_t$, $r_t$, $s_{t+1}$) during its exploring of the maze, and perform gradient descent to train the neural network that indicates how wise an action is for a state. For more information about algorithms and methods used in the agent, please refer to works (Lillicrap et al., 2016; Mnih et al., 2015). Finally, after exploring the maze, the agent is trained, and its knowledge about the maze is stored in the NK-DNA as *Actions*, *States*, *Experiences*, and *Networks*. This allows the agent for sharing and reusing such knowledge in the future.

**CONCLUSIONS AND FUTURE WORK**

In this paper, we proposed the Neural Knowledge DNA, a framework adapting ideas underlying the success of neural networks to knowledge representation for neural network-based knowledge discovering, storing, reusing, improving, and sharing. By taking advantages of neural networks and reinforcement learning, the NK-DNA stores the knowledge learnt through domain's daily operation, and provides an easy way for future accessing, reusing, and sharing such knowledge. At the end of this paper, we tested our proposal idea in an initial experiment, and the results show that the NK-DNA is very promising for knowledge representation, reuse, and sharing among neural network-based AI systems.

For further work, we will do: 1) Refinement and further development of the Deep Learning Engine; 2) Further design and development of the NK-DNA framework, especially, for supporting a range of third-party deep learning frameworks; 3) Exploring and developing methods for reasoning tasks.